%% file: main.tex
\documentclass[runningheads]{llncs}
 
\usepackage{eccv}

\usepackage{eccvabbrv}

\usepackage{graphicx}
\usepackage{booktabs}

\usepackage[accsupp]{axessibility}  
\usepackage{amsmath}
\usepackage{amssymb}
\usepackage{multirow}
\usepackage{bm}
\usepackage{enumitem}
\usepackage{xcolor}
\usepackage{colortbl}
\definecolor{mygray}{gray}{.9}
\usepackage{mathtools}
\usepackage{algorithm}
\usepackage{algpseudocode}
\usepackage{verbatim}
\usepackage{wrapfig}
\usepackage{footnote}

\usepackage{hyperref}
\usepackage{orcidlink}

\begin{document}

\title{GenView: Enhancing View Quality with Pretrained Generative Model for Self-Supervised Learning}

\titlerunning{GenView}



\author{
Xiaojie Li$^{1,2}$\orcidlink{0000-0001-6449-2727} \and 
Yibo Yang$^{3 *}$\orcidlink{0000-0003-0530-7231} \and 
Xiangtai Li$^{4}$\orcidlink{0000-0002-0550-8247} \and 
Jianlong Wu$^{1 *}$\orcidlink{0000-0003-0247-5221} \and 
Yue Yu$^{2}$\orcidlink{0000-0002-9865-2212} \and \\
Bernard Ghanem$^{3}$\orcidlink{0000-0002-5534-587X} \and
Min Zhang$^{1}$\orcidlink{0000-0002-3895-5510}
}

\authorrunning{Li \etal}

\institute{
  \textsuperscript{\rm 1}Harbin Institute of Technology (Shenzhen) \quad
  \textsuperscript{\rm 2}Peng Cheng Laboratory \\ 
  \textsuperscript{\rm 3}King Abdullah University of Science and Technology (KAUST) \quad \\
  \textsuperscript{\rm 4}S-Lab, Nanyang Technological University\\
  {\tt\small \{xiaojieli0903,yibo.yang93,xiangtai94\}@gmail.com, wujianlong@hit.edu.cn, yuy@pcl.ac.cn, bernard.ghanem@kaust.edu.sa,  zhangmin2021@hit.edu.cn} \\
  \url{https://github.com/xiaojieli0903/genview}
}

\maketitle
\let\thefootnote\relax\footnotetext{$*$ Corresponding authors}
\input{sec/0_abstract}
\input{sec/1_intro}
\input{sec/2_relatedwork}
\input{sec/3_method}
\input{sec/4_experiments}
\input{sec/5_conclusion}

\bibliographystyle{splncs04}
\bibliography{main}
\input{sec/X_suppl}
\end{document}

%% file: sec/0_abstract.tex
\vspace{-4mm}
\begin{abstract}
Self-supervised learning has achieved remarkable success in acquiring high-quality representations from unlabeled data. The widely adopted contrastive learning framework aims to learn invariant representations by minimizing the distance between positive views originating from the same image. However, existing techniques to construct positive views highly rely on manual transformations, resulting in limited diversity and potentially false positive pairs. To tackle these challenges, we present GenView, a controllable framework that augments the diversity of positive views leveraging the power of pretrained generative models while preserving semantics. We develop an adaptive view generation method that dynamically adjusts the noise level in sampling to ensure the preservation of essential semantic meaning while introducing variability. Additionally, we introduce a quality-driven contrastive loss, which assesses the quality of positive pairs by considering both foreground similarity and background diversity. This loss prioritizes the high-quality positive pairs we construct while reducing the influence of low-quality pairs, thereby mitigating potential semantic inconsistencies introduced by generative models and aggressive data augmentation. Thanks to the improved positive view quality and the quality-driven contrastive loss, GenView significantly improves self-supervised learning across various tasks. For instance, GenView improves MoCov2 performance by 2.5\%/2.2\% on ImageNet linear/semi-supervised classification. Moreover, GenView even performs much better than naively augmenting the ImageNet dataset with Laion400M or ImageNet21K.
\keywords{Self-supervised learning \and Contrastive learning \and View generation \and Generative models}
\end{abstract}

%% file: sec/1_intro.tex
\vspace{-4mm}
\section{Introduction}
\vspace{-2mm}
\label{sec:intro}

\begin{figure}
 \centering
  \vspace{-3mm}
 \includegraphics[width=0.7\linewidth]{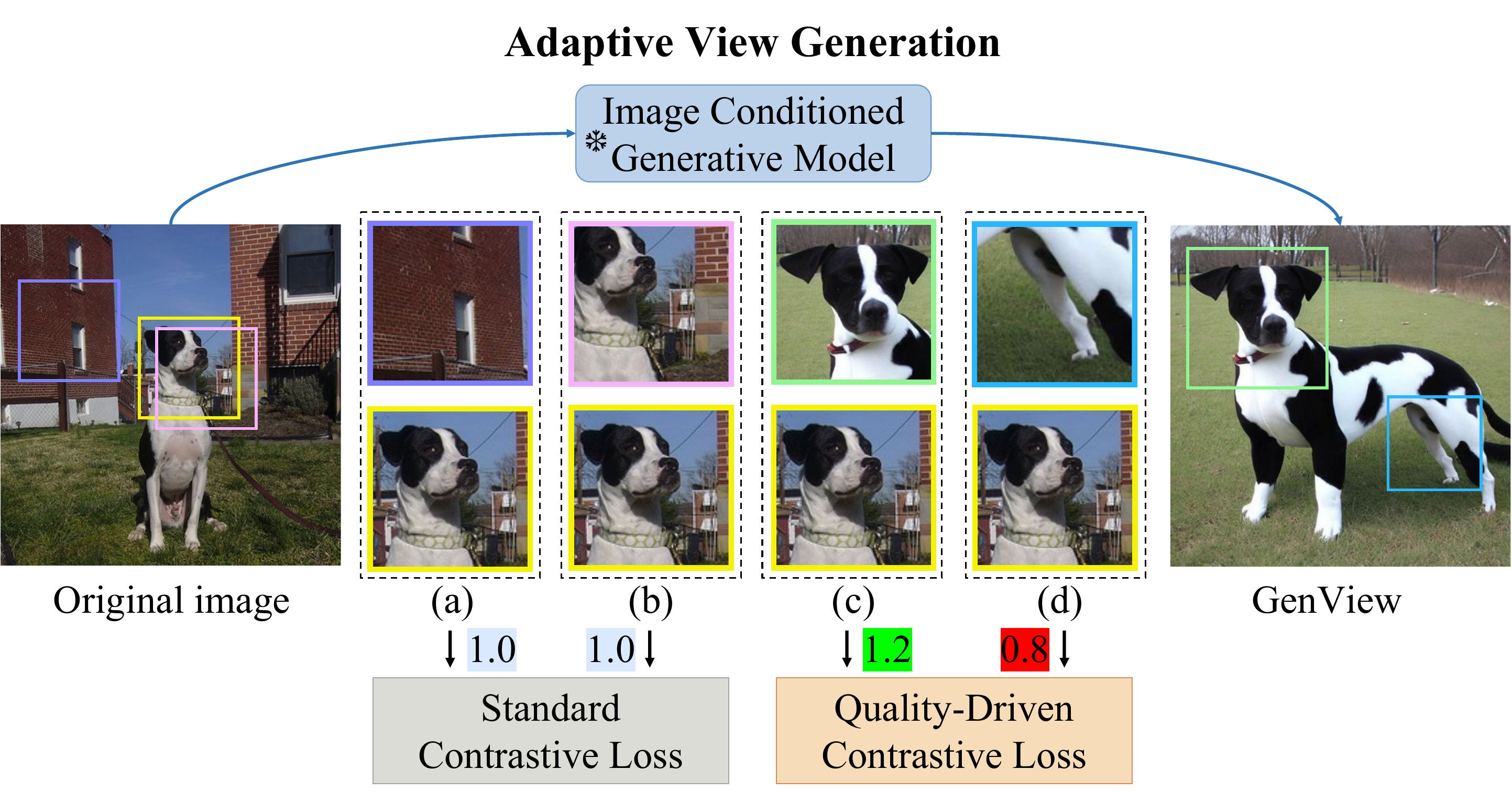}
 \vspace{-3mm}
 \caption{The motivation of GenView: (a) and (b) show standard augmentation-based positive pairs, while (c) and (d) are GenView-constructed pairs. Standard augmentations may cause false positive pair (a) or less diverse pair (b). As a comparison, GenView preserves subject semantics with variations (c and d) and assesses the pair quality to guide contrastive learning. }
\vspace{-6mm}
\label{fig:motivation}
\end{figure}
Self-supervised learning (SSL) has demonstrated remarkable capability in acquiring robust and generalized visual representations from abundant unlabeled data sources \cite{noroozi2016unsupervised, gidaris2018unsupervised, wu2018unsupervised, oord2018representation, he2020momentum, chen2020simple, grill2020bootstrap, zheng2021ressl, huang2022learning, he2022masked, li2023mask,assran2023self,garrido2024learning,wu2019deep,xie2024sscnet,chen2023ccsd,zhang2024twostage}, which can be transferred or leveraged in downstream tasks. Among the various approaches within SSL, Contrastive Learning (CL)~\cite{chen2020simple, chen2020improved, caron2020unsupervised, caron2021emerging, chen2021mocov3, zheng2021ressl, huang2022learning,li2023fine} has emerged as a prominent method, showcasing its effectiveness in numerous downstream tasks (e.g., classification~\cite{he2016deep, li2020local, yang2022towards}, detection~\cite{ren2015faster, li2019detector, he2017mask, carion2020end, wang2022head}, and segmentation~\cite{girshick2014rich, long2015fully,wu2023towards,li2023transformer,OMGSeg}). CL aims to learn invariant representations that remain consistent across various conditions or environments by maximizing the similarity of representations obtained from different distorted versions of a sample, referred to as positive views. Consequently, the construction of high-quality positive views is crucial for CL. A high-quality positive view should retain the semantics of the original images while introducing as much semantic-irrelevant attribute diversity and environmental variations as possible, such that the learned representations can be more generalizable for downstream tasks.

Current CL methods \cite{chen2020simple, grill2020bootstrap, caron2020unsupervised, chen2020improved, caron2021emerging} often employ predefined image augmentations (e.g., random cropping, color distortions, and Gaussian blur) on the same instance to obtain positive views. However, they face two limitations:
\noindent \textbf{(1) Limited Diversity}: Standard augmentations only modify surface-level visual characteristics and fail to introduce new content to capture high-level variations, such as different object viewpoints, textures, or variations within a semantic category. This limitation hinders performance in domains with high intra-category diversity.
\noindent\textbf{(2) False Positive Risk}: Aggressive augmentations are not always precise, potentially leading to false positive pairs. As depicted in \cref{fig:motivation}(a), random cropping of distant patches may miss the entire object, which could mislead the representation learning by minimizing the distance between the object and background in the embedding space. Additionally, as shown in \cref{fig:motivation}(b), cropping nearby patches may fail to introduce sufficient object variations, causing limited diversity in another way.
Advanced methods, such as employing stronger augmentations while preserving task-relevant information~\cite{tian2020makes}, saliency-guided sampling~\cite{selvaraju2021casting}, and center-suppressed sampling~\cite{peng2022crafting}, have been developed to create informative positive pairs. Some methods expand the diversity of positive pairs by utilizing information from the entire training dataset~\cite{caron2020unsupervised, dwibedi2021little}. However, these methods primarily concentrate on optimizing positive views within an instance without introducing new content or incorporating additional information beyond the existing dataset. Consequently, they still have limited ability to capture extensive high-level variations.

Generative models, such as Stable Diffusion \cite{rombach2022high} and DALL-E2 \cite{ramesh2022hierarchical}, have been very successful in generating high-quality diversified images conditioned on an image or embedding. These off-the-shelf pretrained models could help enrich view contents given an image due to their abundant prior knowledge learned from large-scale datasets \cite{schuhmann2022laion, changpinyo2021conceptual}. Albeit they have been leveraged for image classification to address data scarcity~\cite{carlini2023extracting,sariyildiz2023fake,zhou2023training,dunlap2023diversify, trabucco2023effective,burg2023data,zhang2022expanding,ye2023exploiting}, integrating pretrained generative models to pair the images for self-supervised learning is NOT a trivial problem. Despite the strong generative ability, these models may be pretrained on the datasets from different distributions, and the sampling process is not determinant. As a result, they will still inevitably face the risk of generating images with different semantics from the conditional images, resulting in false positive pairs. This presents a key challenge: how to appropriately control the randomness of generation while maintaining semantic consistency to help SSL in a controllable way.

To address these challenges, we introduce \textbf{GenView}, a controllable framework that enhances view quality for SSL using the powerful pretrained generative model, and guide contrastive learning via quality assessment. In our framework, as shown in \cref{fig:motivation}, given an image as the source view, we construct its positive view using the synthetic image sampled from a pretrained generative model conditioned on this image. To optimally balance the trade-off between diversity and semantic fidelity, we develop an adaptive view generation method, which dynamically adjusts the noise level of the generative model to control the extent of perturbation applied to the conditional image embedding. We calculate the proportion of the foreground area within an input image. If the subject is not prominent with a low foreground proportion, it reduces the perturbation strength to ensure the correct semantic content of the synthetic image. If the subject is clear and distinguishable with a high foreground proportion, it increases the perturbation strength to create more variations for more diverse content and environments. As depicted in \cref{fig:motivation}(c), the view constructed by our method has a different pose and environment compared to the traditional way.

Even with our adaptive view generation, false positive pairs are still inevitable because both the sampling of the generative model and cropping are not determinant. To further mitigate the effect of potential false positive pairs that could mislead contrastive learning, we introduce a quality-driven contrastive loss to guide the contrastive learning with pair quality. Concretely, we assess the quality of positive pairs considering both foreground similarity and background diversity. It prioritizes the positive pairs with high foreground similarity to ensure semantic coherence, while also favoring the pairs with low background similarity to promote diverse environments for learning invariant representations. We then recalibrate the contrastive loss function by reweighting each pair with its pair quality, which enhances the contributions of high-quality positive pairs, and simultaneously reduces the influence of low-quality and even false pairs. As illustrated in \cref{fig:motivation}(c) and (d), our quality-driven contrastive loss assigns a higher score to the high-quality positive pair and a lower score to the pair with a relatively lower quality. In summary, the contributions of this paper include:
\vspace{-1mm}
\begin{itemize}
  \item We introduce GenView framework, which enhances the view quality for SSL leveraging the power of pretrained generative model in a controllable way. An adaptive view generation method is developed to construct positive views, balancing the trade-off between diversity and semantic fidelity. 
  \item We propose a quality-driven contrastive loss that prioritizes high-quality positive pairs to guide the contrastive learning with pair quality, further mitigating the impact of low-quality and false pairs.
  \item In experiments, GenView significantly enhances the performance of popular contrastive learning algorithms including MoCov2 \cite{chen2020improved}, SimSiam \cite{chen2020exploring}, BYOL \cite{grill2020bootstrap}, and MoCov3 \cite{chen2021mocov3} on various downstream tasks such as linear/semi-supervised classification, semantic segmentation, and object detection. Particularly, GenView also performs better than naively augmenting the ImageNet1K dataset with Laion400M or ImageNet21K.
\end{itemize}

%% file: sec/2_relatedwork.tex
\vspace{-5mm}
\section{Related Work}
\vspace{-2.5mm}
\paragraph{\textbf{Self-Supervised Learning.}}
Self-supervised learning is a promising paradigm for representation learning, relying on unlabeled data and pretext tasks such as auto-encoders~\cite{pathak2016context, vincent2008extracting}, image pixel generation~\cite{goodfellow2014generative, kingma2014auto}, rotation prediction~\cite{gidaris2018unsupervised}, jigsaw puzzles~\cite{noroozi2016unsupervised}, and mask image modeling~\cite{bao2021beit, he2022masked}. In recent years, contrastive learning (CL) methods~\cite{wu2018unsupervised, oord2018representation, tian2019contrastive, he2020momentum, chen2020simple, chen2020improved, zheng2021ressl, huang2022contrastive} have significantly improved SSL by reducing the distance between representations of positive pairs and increasing the distance between representations of negative pairs in the latent feature space simultaneously. Complementing CL approaches, various non-CL methods have emerged, seeking alternatives to negative samples and strategies to prevent network output collapse \cite{caron2018deep, asano2020self, li2020prototypical, caron2020unsupervised, chen2020exploring, grill2020bootstrap, zbontar2021barlow, ermolov2021whitening, caron2021emerging}.

The construction of a pair of views is crucial in contrastive learning \cite{tian2019contrastive, chen2020simple, caron2020unsupervised}, and traditional SSL generates positive views through hand-designed augmentations, which may face limited diversity and induce semantically irrelevant pairs. Later studies introduce stronger augmentations preserving task-relevant information~\cite{tian2020makes}, unsupervised saliency maps for cropping constraints~\cite{selvaraju2021casting}, and center-suppressed sampling for increased diversity~\cite{peng2022crafting}. Clustering-based methods~\cite{caron2018deep, caron2020unsupervised} and neighborhood-based methods~\cite{dwibedi2021little} expand the diversity of positive pairs by leveraging information from the training dataset. However, the diversity introduced is ultimately confined to the scope of the training dataset, limiting the ability to capture extensive diversity for learning more generalizable representation. In our method, we break free from this limitation by utilizing the pretrained image-conditioned generative model for high-quality view generation. 

\vspace{-2.5mm}
\paragraph{\textbf{Generative Models.}}
Various generative models, including VAEs~\cite{kingma2014auto,razavi2019generating}, GANs~\cite{goodfellow2014generative,karras2020training,brock2018large,li2022eliminating}, autoregressive models~\cite{ramesh2021zero}, and diffusion models~\cite{ho2020denoising,ho2022classifier,rombach2022high,ramesh2022hierarchical,bie2023renaissance,wu2024lgvi} (DMs), have demonstrated the ability to create highly realistic images. Particularly, DMs such as Imagen~\cite{saharia2022photorealistic}, GLIDE~\cite{nichol2022glide}, Stable Diffusion~\cite{rombach2022high}, and DALL-E2~\cite{ramesh2022hierarchical}, trained on extensive large-scale datasets such as LAION-5B~\cite{schuhmann2022laion} and CC12M~\cite{changpinyo2021conceptual}, have excelled in generating photorealism images.
Recent research has explored generative models for data augmentation in various tasks, including classification~\cite{he2022synthetic,li2023synthetic,shipard2023diversity,dunlap2023diversify,ye2023exploiting,luo2023rethinking,sariyildiz2023fake}, segmentation~\cite{li2023guiding,xie2023mosaicfusion, wang2022art,wu2023towards,li2023transformer}, and test-time optimization~\cite{feng2023diverse}. In representation learning, GANs~\cite{tamkin2020viewmaker}, instance-conditioned GANs~\cite{astolfi2023instance,yang2022local}, neural transformation networks~\cite{kim2023neural}, and DMs~\cite{zang2023boosting} have been employed to introduce more variations. However, the diversity introduced is still constrained by the training dataset used for SSL.

Instead of training generative models from scratch, some methods use pretrained generative models to augment representation learning, leveraging the prior knowledge learned from large-scale datasets \cite{schuhmann2022laion, changpinyo2021conceptual} to enhance the high-level diversity of the generated views \cite{jahanian2021generative, he2022synthetic, tian2023stablerep, dunlap2023diversify, shipard2023diversity, trabucco2023effective,zhang2023free}. However, these models rely on \textit{constant} \cite{he2022synthetic, dunlap2023diversify, tian2023stablerep, zhang2023free} or \textit{random} \cite{shipard2023diversity, trabucco2023effective, zhou2023training} hyperparameters to determine the extent of deviation in the generated images. This can lead to uncontrolled data generation characterized by inconsistent semantics with the conditional image, reducing the quality of positive pairs. In contrast, our approach employs adaptive view generation that controls the noise level when sampling images to keep a balance between semantic fidelity and diversity based on individual image characteristics. We also propose a quality-driven contrastive loss to enhance the contributions of high-quality positive pairs while diminishing the impact of low-quality and false pairs.

%% file: sec/3_method.tex
\vspace{-4mm}
\section{Method}
\label{method}
\vspace{-2mm}
In this section, we first provide a review of self-supervised learning in \cref{method:ssl}. We introduce our framework in \cref{method:framework}. Then, we develop adaptive view generation and quality-driven contrastive loss in \cref{method:adaptive} and \ref{method:quality}. 

\begin{figure*}[!t]
  \centering
  \includegraphics[width=1\linewidth]{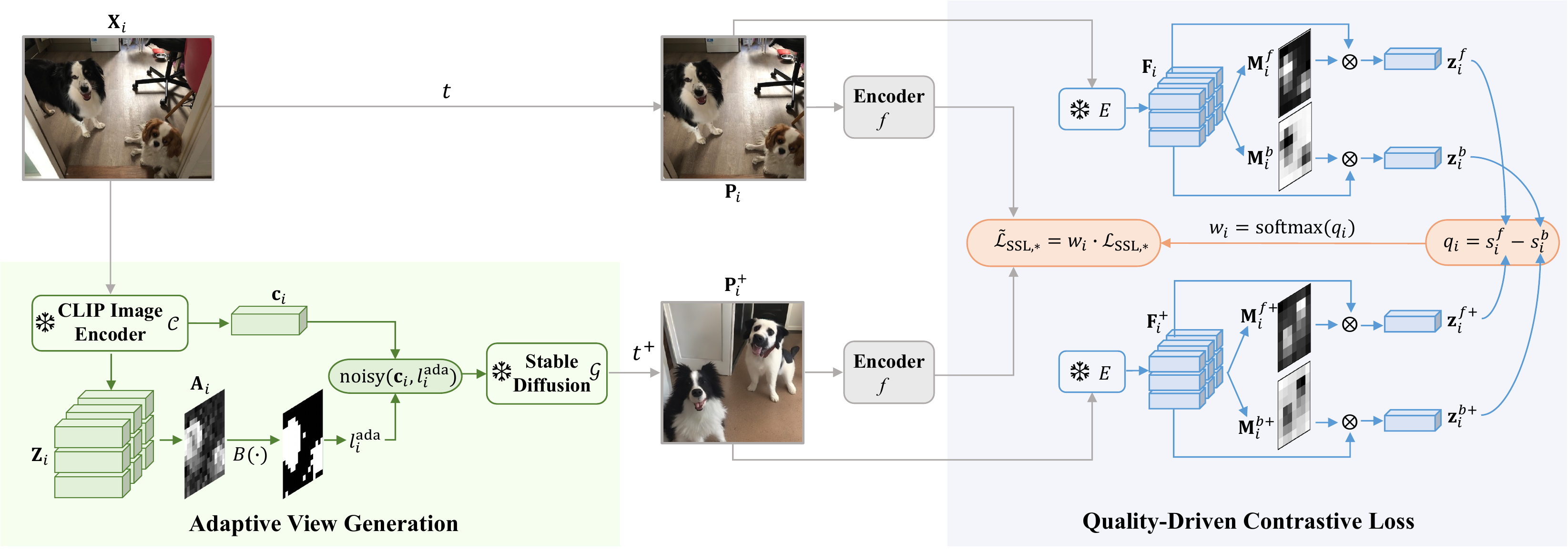}
  \vspace{-5mm}
  \caption{GenView is composed of a view quality enhancement framework, an adaptive view generation method to balance diversity and semantic fidelity, and a quality-driven contrastive loss mechanism. The framework generates the enhanced view by passing the noisy image embedding, which is extracted from the frozen CLIP encoder, to the image-conditioned pretrained generative models (the Stable Diffusion generator). Positive views are passed through encoders to compute the contrastive loss, with an emphasis on those high-quality positive pairs. The encoders $f$ can be the same encoder or different ones, \eg an encoder and its momentum-updated one. All the pretrained CLIP encoder and Stable Diffusion have not accessed the dataset for SSL.}
  \label{fig:framework}
  \vspace{-6mm}
\end{figure*}

\vspace{-3mm}
\subsection{Preliminaries on Self-Supervised Learning}
\label{method:ssl}
\vspace{-1mm}
Current SSL frameworks often create positive pairs ($\mathbf{P}_i^1, \mathbf{P}_i^2$) for each instance $\mathbf{X}_i$ in a batch of $n$ images $\mathbf{X}_{1:n}=\{\textbf{X}_i\}^n_{i=1}$. These pairs are generated by applying random predefined augmentations to the same instance:
\begin{equation}\label{eq:normal_aug}
\textbf{P}_i^1 = t^1(\mathbf{X}_i), \quad \textbf{P}_i^2 = t^2(\mathbf{X}_i),
\end{equation}
where the augmentations, $t^1(\cdot)$ and $t^2(\cdot)$, can either be from the same ($t^1, t^2 \sim \mathcal{T}$) or from different distributions ($t^1 \sim \mathcal{T}$, $t^2 \sim \mathcal{T}'$).
The encoder network $f(\cdot)$ is then applied to $\mathbf{P}_i^1$ to extract the representation, resulting in $\mathbf{h}^1_i = f(\textbf{P}_i^1)$. These representations are projected into an embedding space using a two-layer non-linear projection head, denoted as $\mathbf{z}^1_i = g(\mathbf{h}^1_i)$. Additionally, $\mathbf{P}_i^2$ can be encoded using the same encoder and projection head as $\mathbf{P}_i^1$~\cite{chen2020exploring,caron2020unsupervised}, or their momentum-updated versions~\cite{he2020momentum,grill2020bootstrap}.

Various SSL frameworks, including SimCLR \cite{chen2020simple} and MoCo \cite{he2020momentum}, use the noise contrastive estimation objective $\mathcal{L}_{\text{SSL, NCE}}$ to distinguish between instances:
\begin{equation}\label{eq:loss_ssl_nce}
\mathcal{L}_\text{SSL, NCE} = -\log \frac{\exp(\mathbf{z}_i^1 \cdot \mathbf{z}^2_i/ \tau) }{ \exp(\mathbf{z}_{i}^1 \cdot \mathbf{z}_{i}^2 / \tau ) + \sum_{k=1}^{N} \exp(\mathbf{z}_{i}^1 \cdot \mathbf{z}_{k} / \tau ) },
\end{equation}
with $\tau$ as the temperature parameter. Additionally, methods like BYOL \cite{grill2020bootstrap} and SimSiam \cite{chen2020exploring} introduce a non-linear predictor head $q(\cdot)$ to map $\mathbf{z}$ to $\mathbf{p}$, minimizing negative cosine similarity $\mathcal{L}_{\text{SSL, COS}}$ as:
\begin{equation}\label{eq:loss_ssl_cos}
\mathcal{L}_\text{SSL, COS} = - \frac{\mathbf{p}_i^1}{\lVert \mathbf{p}_i^1 \rVert} \cdot \frac{\mathbf{z}_i^2}{\lVert \mathbf{z}_i^2 \rVert}.
\end{equation}

SwAV \cite{caron2020unsupervised} employs a linear mapping of positive embeddings $\mathbf{z}^1$ and $\mathbf{z}^2$ to learned prototypes to obtain ``codes'' $\tilde{\mathbf{z}}^1$ and $\tilde{\mathbf{z}}^2$. The targets are transformed with a Sinkhorn-Knopp ($SK$) step. Then the Kullback-Leibler divergence loss $\mathcal{L}_{\text{SSL, KL}}$ is computed as:
\begin{equation}
\mathcal{L}_\text{SSL, KL} = D_\text{KL}( \tilde{\mathbf{z}}^1 \| SK(\tilde{\mathbf{z}}^2) ).
\label{eq:loss_ssl_kl}
\end{equation}

In experiments, we will integrate GenView on all these popular SSL methods to test its generalizability. 

\vspace{-4mm}
\subsection{Our Framework}
\vspace{-2mm}
\label{method:framework}
The framework of our method is depicted in \cref{fig:framework}. Traditional methods face the challenge of limited view diversity by generating positive pairs by applying augmentation twice to the same instance, as illustrated in \cref{eq:normal_aug}. To this end, we employ an image-conditioned pretrained generative model to enhance the view quality. Specifically, we utilize the Stable unCLIP model, an extension of Stable Diffusion~\cite{rombach2022high} with unCLIP~\cite{ramesh2022hierarchical}, fine-tuned to accept CLIP~\cite{radford2021learning} ViT-H/14 image embeddings in addition to text encodings. To improve the diversity of positive views, we inject Gaussian noise perturbations to the conditional image embedding through a diffusion process $\textbf{noisy}(\cdot, l)$, which adds $l$ steps of Gaussian noise to the conditional image embedding. The degree of variation in the final images is controlled by the perturbation strength $l$, with a higher value leading to an increased diversity.

The generation stage starts with a random normal distribution $\mathbf{z}_T \sim \mathcal{N}(0, \mathbf{I})$, where $T$ represents the denoising steps of the generation process. The pretrained diffusion model $\mathcal{G}(\cdot)$, conditioned on the noisy image embeddings, iteratively denoises the latent features. The synthetic positive view can be defined as: 
\begin{equation}
    \mathbf{X}^+_i = \mathcal{G}(\mathbf{z}_T, \textbf{noisy}(\textbf{c}_i, l), w),
\label{eq:x_constant}
\end{equation}
where $w$ refers to the pretrained parameters of the generative model, and $\mathbf{c}_i$ represents the conditional image embedding obtained from the CLIP image encoder as $\mathbf{c}_i = C(\textbf{X}_i)$. 

We then design a pair construction mechanism by leveraging the original image as one view and pairing it with another view generated by the generative model for contrastive learning.
Specifically, hand-designed data augmentations ($t \sim \mathcal{T}$ for the original image and $t^{+} \sim \mathcal{T}$ or $\mathcal{T}'$ for the synthetic image) are applied to create an enhanced pair of positive views $(\mathbf{P}_i, \mathbf{P}_i^+)$:
\begin{equation}
  \mathbf{P}_i = t(\mathbf{X}_i), \quad \mathbf{P}_i^+ = t^{+}(\mathbf{X}_i^+).
\end{equation}

Through this mechanism, we significantly increase view diversity by leveraging the capabilities of the generative model, as illustrated in \cref{fig:motivation}. Meanwhile, unlike most generative model-based augmentation methods~\cite{tamkin2020viewmaker, kim2023neural, yang2022local}, which generate positive pairs from two synthetic images derived from the same original image, GenView integrates the original image itself as one of the views. This approach effectively controls potential feature drift caused by domain differences between the dataset used to train the generative model and the current pre-training dataset. Furthermore, when the synthetic image contains noise, such as artifacts or semantic discrepancies, the presence of the original real image prevents excessive deviation in feature learning. Thus, while enhancing the view diversity, our framework maintains stability and fidelity when combining the traditional augmentation with the strength of the generative model.

\vspace{-4mm}
\subsection{Adaptive View Generation}
\begin{figure}[t]
\centering
\includegraphics[width=0.7\linewidth]{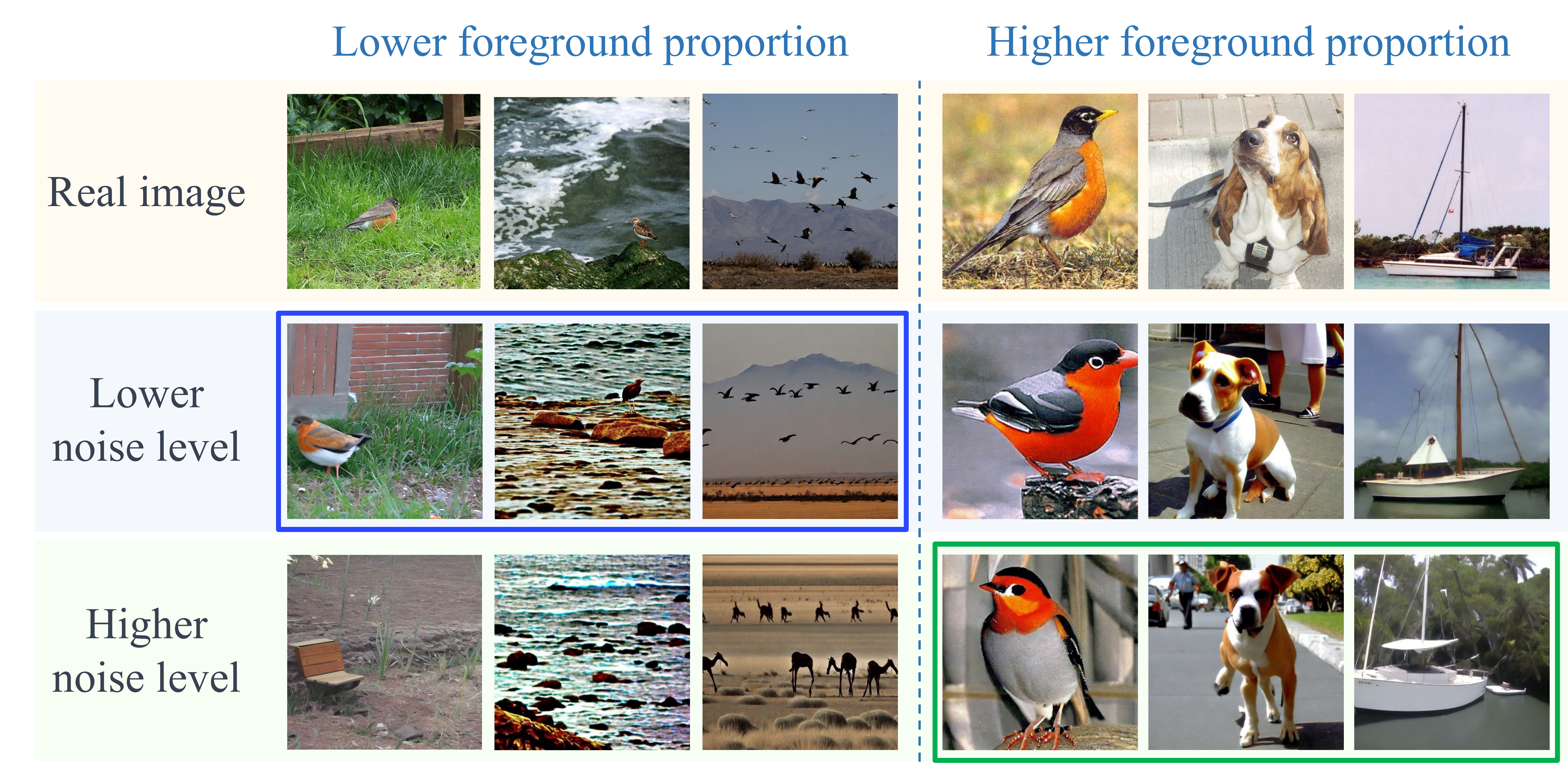}
\vspace{-3mm}
\caption{Illustration of our adaptive view generation. For the images with lower foreground proportion, a lower noise level is selected (in blue) because a higher noise level could easily result in synthetic images whose semantic contents are changed (1st column), disappeared (2nd column), or distorted (3rd column). For the images with higher foreground proportion, a higher noise level is favored  (in green) to introduce diversity, \emph{e.g.} different pose (4th column), action (5th column), and background (6th column).
}
\vspace{-5mm}
\label{fig:adaptive_noise_level}
\end{figure}

\vspace{-2mm}
\label{method:adaptive}
To address the concerns related to inappropriate noise levels during image generation, we develop an adaptive view generation method, which dynamically adjusts the noise level based on the proportion of the foreground content. This introduces diverse positive pairs while ensuring coherent subject semantics. Given a conditional image $\mathbf{X}_i$, we employ a pretrained CLIP image encoder $C(\cdot)$ to extract latent features $\mathbf{Z}_i \in \mathbb{R}^{H \times W \times K}$, where $H$, $W$, and $K$ represent the height, width, and the dimension of features, respectively. To separate the image's main object from the background, we perform Principal Component Analysis (PCA) among features for all images and obtain the first component. Then, we apply min-max normalization to generate attention maps $\mathbf{A}_i \in \mathbb{R}^{H \times W}$, where higher values indicate a higher probability of being foreground content. The proportion of foreground content, denoted as ${p}_i$, is calculated as follows:
\begin{equation}
\label{eq:proportion}
{p}_i = \frac{\sum_{h=1}^H \sum_{w=1}^W B(\mathbf{A}_{i,h,w}, a)}{H \times W},
\end{equation}
where $B(\cdot, a)$ represents a binary thresholding function with $a$ as the threshold. To map the proportion to the noise level $l^\text{ada}$, we introduce a function $\mathcal{F}^{\text{ada}}$. The range of the ratio ${p}$ is evenly divided into 5 intervals, and values are mapped to discrete scales: $\{0, 100, 200, 300, 400\}$. To reduce the risk of excessive distortion from higher noise levels, we limit the maximum at 400, even though noise levels during training could reach up to 1000. The adaptive noise level $l^\text{ada}_i$ is calculated as follows:
\begin{equation}
l^\text{ada}_i = \mathcal{F}^{\text{ada}}({p}_i) = 100 \cdot \left\lfloor \frac{{p}_i}{0.2} \right\rfloor,
\label{eq:l_ada}
\end{equation}
where $\left\lfloor \cdot \right\rfloor$ rounds down to the nearest integer. Our approach adapts noise levels to the characteristics of images, and thus effectively balances the trade-off between semantic fidelity and diversity in generated images. As illustrated in \cref{fig:adaptive_noise_level}, the selected noise level (in blue) is low for the images with a lower foreground proportion to better preserve their semantic contents, while for those with a higher proportion, a high noise level is adopted (in green) to introduce more diversity because the key subjects are less likely to be changed or disappeared in their generated images. The adaptive generated positive view is defined as: 
\begin{equation}
    \mathbf{X}^+_i = \mathcal{G}(\mathbf{z}_T, \textbf{noisy}(\textbf{c}_i, l_i^\text{ada}), w).
\label{eq:x_ada}
\end{equation}

This process works in an offline manner before SSL training, so does not increase the burden on training time. Besides, the offline view generation is once-for-all and the generation result can be re-used multiple times for various baselines.

 \vspace{-2mm}
\subsection{Quality-driven Contrastive Loss}
\vspace{-2mm}
\label{method:quality}
In this section, we introduce a quality-driven contrastive loss that guides contrastive learning by assessing the quality of positive pairs. It prioritizes the pairs with high foreground similarity and low background similarity to facilitate the learning of invariant representations.

Given a pair of positive views $(\mathbf{P}_i, \mathbf{P}^+_i)$, we employ a frozen encoder that is pretrained by CLIP (without accessing the dataset for SSL), denoted as $E(\cdot)$, to extract feature maps $\mathbf{F}_i, \mathbf{F}_i^+ \in \mathbb{R}^{H' \times W' \times K'}$. PCA is performed on feature maps, and min-max normalization is applied to the first component of PCA features, generating foreground attention maps $\mathbf{M}_i^f, \mathbf{M}_i^{f+} \in \mathbb{R}^{H' \times W'}$. The background activation map for the $i$-th sample is defined as $\mathbf{M}^b_i = 1 - \mathbf{M}^f_i$. Subsequently, we use these maps to aggregate feature maps into foreground and background representations, yielding $\mathbf{z}^{f}_{i},\mathbf{z}^{f+}_{i}, \mathbf{z}^{b}_{i},\mathbf{z}^{b+}_{i} \in \mathbb{R}^{K'}$, which can be computed as follows:
\begin{equation}
\begin{aligned}
    \mathbf{z}^{f}_i &= \mathbf{M}_i^f \otimes \mathbf{F}_i, \quad \mathbf{z}^{f+}_i = \mathbf{M}_i^{f+} \otimes \mathbf{F}_i^{+}, \\
    \mathbf{z}^{b}_i &= \mathbf{M}_i^b \otimes \mathbf{F}_i, \quad \mathbf{z}^{b+}_i = \mathbf{M}_i^{b+} \otimes \mathbf{F}_i^{+},
\end{aligned}
\end{equation}
where the operation $\otimes$ represents spatial aggregation defined as $\mathbf{z} = \mathbf{M} \otimes \mathbf{F} = \sum_{h=1}^{H} \sum_{w=1}^{W} \mathbf{M}_{h, w}\mathbf{F}_{h, w, *}$. We calculate the foreground-foreground similarity $s^f_i$ and background-background similarity $s^b_i$ as follows:
\begin{equation}
  s^f_i = \text{sim}(\mathbf{z}^f_i, \mathbf{z}^{f+}_i), \quad s^b_i = \text{sim}(\mathbf{z}^b_i, \mathbf{z}^{b+}_i),
\end{equation}
where $\text{sim}(\cdot, \cdot)$ denotes the cosine similarity of the input representations. Next, we introduce a quality score for each positive pair:
\begin{equation}
  q_i = s_i^f - s_i^b.
\end{equation}

We then propose a re-weighting factor denoted as $w_i$, based on the computed pair qualities of a batch of images, to adjust the contribution of each pair to the overall loss during contrastive training:
\begin{equation}
  w_i = \frac{\text{exp}(q_i)}{\sum_{j=1}^n \text{exp}(q_j)}.
\end{equation}

The re-weighting factor $w_i$ is used to balance the influence of different pairs, allowing us to prioritize the pairs with higher foreground similarity and lower background similarity, and also mitigate the potential influence of those low-quality or wrong positive pairs. The final contrastive loss is defined as:
\begin{equation}
\tilde{\mathcal{L}}_\text{SSL,*} = w_i \mathcal{L}_\text{SSL,*},
\label{eq:loss_quality}
\end{equation}
where $\mathcal{L}_\text{SSL,*}$ can be any contrastive loss in Eqs. (\ref{eq:loss_ssl_nce})-(\ref{eq:loss_ssl_kl}).

%% file: sec/4_experiments.tex
\section{Experiments}
\vspace{-2mm}
\label{sec:exp}
We compare GenView with state-of-the-art SSL methods, including MoCov2~\cite{he2020momentum}, BYOL~\cite{grill2020bootstrap}, SwAV~\cite{caron2020unsupervised}, SimSiam~\cite{chen2020exploring}, and MoCov3~\cite{chen2021mocov3}. We experiment with various network architectures, such as ResNet-18~\cite{he2016deep}, ResNet-50~\cite{he2016deep}, ViT-S~\cite{dosovitskiy2020image}, and ViT-B~\cite{dosovitskiy2020image}. By default, ResNet-50 serves as the backbone. ViT-S and ViT-B are adopted for comparison with MoCov3. For details on adaptive view generation and quality-driven contrastive loss implementations for different pretraining datasets, please refer to the \cref{supp_details,supp_algo}.

\subsection{Main Results}
\label{exp:main}
\noindent\textbf{\textit{Linear classification.}}

\begin{wraptable}{r}{0.5\columnwidth}
\vspace{-15mm}
\begin{center}
 \centering
 \small
 \caption{\textbf{Linear evaluation on IN-1K}. $*$: our reproduction.}
 \scalebox{0.8}{
 \begin{tabular}{lccc}
  \toprule
    Method & Architecture & Epochs & Top-1 \\
    \midrule
    InstDisc~\cite{wu2018unsupervised} & ResNet-50 & 200 & 56.5 \\
    SimCLR~\cite{chen2020simple} & ResNet-50 & 200 & 66.8 \\
    PCL~\cite{li2020prototypical} & ResNet-50 & 200 & 67.6 \\
    Adco~\cite{qi2020learning} & ResNet-50 & 200 & 68.6 \\
    InfoMin~\cite{tian2020makes} & ResNet-50 & 200 & 70.1\\
    NNCLR~\cite{dwibedi2021little} & ResNet-50 & 200 & 70.7 \\
    LEVEL~\cite{huang2022learning} & ResNet-50 & 200 & 72.8 \\
    Barlow Twins~\cite{zbontar2021barlow} & ResNet-50 & 300 & 71.4 \\
    CLIP~\cite{radford2021learning} & ResNet-50 & - & 74.3 \\
    \midrule
    MoCov2~\cite{he2020momentum} & ResNet-50 & 200 & 67.5 \\ 
    MoCov2 + C-Crop~\cite{peng2022crafting} & ResNet-50 & 200 & 67.8\\
    \rowcolor{mygray} MoCov2 + GenView & ResNet-50 & 200 & \textbf{70.0} \\
    SwAV~\cite{caron2020unsupervised}$^*$& ResNet-50 & 200 & 70.5 \\
    \rowcolor{mygray} SwAV + GenView& ResNet-50& 200&\textbf{71.7}\\
    SimSiam~\cite{chen2020exploring} & ResNet-50 & 200 & 70.0\\
    \rowcolor{mygray} SimSiam + GenView & ResNet-50 & 200 & \textbf{72.2} \\
    BYOL~\cite{grill2020bootstrap}$^*$ & ResNet-50 & 200 & 71.8 \\
    \rowcolor{mygray} BYOL + GenView & ResNet-50 & 200 & \textbf{73.2} \\ 
    MoCov3~\cite{chen2021mocov3}& ResNet-50& 100 & 68.9 \\ 
    \rowcolor{mygray} MoCov3 + GenView & ResNet-50 & 100 & \textbf{72.7} \\
    MoCov3~\cite{chen2021mocov3}& ResNet-50& 300&72.8\\ 
    \rowcolor{mygray} MoCov3 + GenView & ResNet-50 & 300& \textbf{74.8}\\
    \midrule
    MoCov3~\cite{chen2021mocov3} & ViT-S & 300 & 73.2\\ 
    \rowcolor{mygray} MoCov3 + GenView & ViT-S & 300 & \textbf{74.5}\\
    MoCov3~\cite{chen2021mocov3} & ViT-B & 300 & 76.7 \\ 
    \rowcolor{mygray} MoCov3 + GenView & ViT-B & 300 & \textbf{77.8} \\
  \bottomrule
\label{tab:exp_linear}
 \end{tabular}
 }
\end{center}
\vspace{-15mm}
\end{wraptable}
\noindent GenView is framework-agnostic, allowing flexibility with SSL frameworks and associated training components like backbone networks, loss functions, and optimizers. To ensure fair comparisons, we maintain consistent pretraining settings as baseline methods on ImageNet-1K~\cite{deng2009imagenet} (IN-1K). To evaluate our method, we follow a standard linear classification protocol, as described in previous works~\cite{chen2020simple,chen2020improved,grill2020bootstrap}. The linear classifier is trained on top of the frozen representation for 90 epochs with a batch size of 1,024, an initial learning rate of 0.4, an SGD optimizer with 0.9 momentum and no weight decay, and the cosine-annealed learning rate schedule~\cite{loshchilov2016sgdr}. For ViT-based models, the initial learning rate is set to 12.
\cref{tab:exp_linear} presents the results of top-1 accuracy on the validation set of IN-1K.
GenView consistently improves SSL performance across various frameworks, including ResNet-50 and Transformer architectures like ViT-S and ViT-B. Its effectiveness is maintained across different pretraining epochs, outperforming the MoCov3 baselines pretrained for 100 or 300 epochs. GenView outperforms C-Crop~\cite{peng2022crafting} that also constructs better views, highlighting our advantage in utilizing pretrained generative models' prior knowledge to create diverse views in a controlled manner. GenView can complement both contrastive (\eg MoCov2 and MoCov3) and non-contrastive methods (\eg BYOL, SimSiam, and SwAV), addressing their limitations of positive pair quality. These results demonstrate GenView's consistent ability in enhancing the linear classification performance of various SSL models. It's noted that when GenView is integrated with MoCov3 utilizing a ResNet-50 backbone and pretrained over 300 epochs, it achieves competitive performance (74.8\% with 1.28 million images) compared to CLIP (74.3\% on WebImageText with 400 million pairs), highlighting GenView's efficiency.

\begin{table}[!t]
    \begin{minipage}{0.45\linewidth}
    \centering
      \small
      \caption{\textbf{Comparison with existing SSL methods for semi-supervised learning on IN-1K}. Models with ResNet-50 backbone are pretrained on IN-1K. $*$: our reproduction.}
      \vspace{-3mm}
      \label{tab:exp_semi}
      \scalebox{0.7}{
      \begin{tabular}{@{}lccccc@{}}
        \toprule
        \multirow{2}{*}{Method} & \multirow{2}{*}{Epochs}&\multicolumn{2}{c}{1\% Labels} & \multicolumn{2}{c}{10\% Labels} \\
        \cmidrule(l{3pt}r{3pt}){3-4} \cmidrule(l{3pt}r{3pt}){5-6} 
         & &Top-1 & Top-5 & Top-1 & Top-5 \\
        \midrule
        PCL~\cite{li2020prototypical} & 200&- & 75.6 & - & 86.2 \\
        SwAV~\cite{caron2020unsupervised} & 800 & 53.9 & 78.5 & 70.2 & 89.9 \\
        SimCLR~\cite{chen2020simple} & 1000& 48.3 & 75.5 & 65.6 &87.8 \\ 
        Barlow Twins~\cite{zbontar2021barlow} & 1000 & 55.0 & 79.2 & 69.7 & 89.3 \\
        NNCLR~\cite{dwibedi2021little} & 1000 & 56.4 & 80.7 & 69.8 & 89.3 \\ 
        \midrule
        MoCov3~\cite{chen2021mocov3}$^*$& 100& 50.4& 76.6& 66.8&88.4\\
        \rowcolor{mygray} MoCov3 + GenView& 100& \textbf{51.9}& \textbf{78.5}&\textbf{ 68.4}&\textbf{89.4}\\
        MoCov2~\cite{he2020momentum}$^*$ & 200 &42.1& 70.9& 60.9& 84.2\\
        \rowcolor{mygray} MoCov2 + GenView & 200 &\textbf{50.6} & \textbf{78.3} & \textbf{63.1} &\textbf{ 86.0} \\
        BYOL~\cite{grill2020bootstrap}$^*$ & 200 &53.2& 78.8& 68.2& 89.0\\
        \rowcolor{mygray} BYOL + GenView & 200 &\textbf{55.6} & \textbf{81.3} & \textbf{68.6} & \textbf{89.5} \\
        MoCov3~\cite{chen2021mocov3}$^*$& 300& 56.2& 80.7& 69.4&89.7\\
        \rowcolor{mygray} MoCov3 + GenView& 300& \textbf{58.1}& \textbf{82.5}& \textbf{70.6}&\textbf{90.4}\\
        \bottomrule
      \end{tabular}
      }
     
  \end{minipage}
  \hfill
  \begin{minipage}{0.5\linewidth} 
      \centering
     \small
     \caption{\textbf{Transfer learning on MS-COCO object detection and instance segmentation}. Models with ResNet-50 backbone are pretrained for 200 epochs on IN-1K. $*$: our reproduction.}
     \vspace{3mm}
 \label{tab:exp_coco}
 \scalebox{0.7}{
 \begin{tabular}{@{}lcccccc@{}}
  \toprule
  \multirow{2}{*}{Method} & \multicolumn{3}{c}{Object Det.} & \multicolumn{3}{c}{Instance Seg.} \\
   \cmidrule(l{3pt}r{3pt}){2-4} \cmidrule(l{3pt}r{3pt}){5-7}
   & AP & AP$_{50}$ & AP$_{75}$ & AP & AP$_{50}$ & AP$_{75}$ \\
  \midrule
  ReSim~\cite{xiao2021region} & 39.8 & 60.2 & 43.5 & 36.0 & 57.1 & 38.6 \\
  DenseCL~\cite{wang2021dense} & 40.3 & 59.9 & 44.3 & 36.4 & 57.0 & 39.2 \\ \midrule
  SimSiam~\cite{chen2020exploring}$^*$& 38.5 & 57.8 & 42.3 & 34.7 & 54.9 & 37.1 \\
  \rowcolor{mygray} SimSiam + GenView& \textbf{39.1} & \textbf{58.5} & \textbf{43.0} & \textbf{35.2}& \textbf{55.9} & \textbf{37.7}\\ \midrule
  MoCov2~\cite{chen2020improved}$^*$ & 39.7 & 59.4 & 43.6 & 35.8 & 56.5 & 38.4 \\ 
  MoCov2 + FreeATM~\cite{zhang2023free} & 40.1& -& -& -& -&- \\ 
  \rowcolor{mygray} MoCov2 + GenView & \textbf{40.5} & \textbf{60.0} & \textbf{44.3} & \textbf{36.3} & \textbf{57.1} & \textbf{38.9} \\\midrule
  BYOL~\cite{grill2020bootstrap}$^*$ & 40.6 & 60.9 & 44.5 & 36.7 & 58.0 & 39.4 \\
  \rowcolor{mygray} BYOL + GenView & \textbf{41.2} & \textbf{61.5} & \textbf{44.9} & \textbf{37.0} & \textbf{58.4} & \textbf{39.7} \\
  \bottomrule
 \end{tabular}
     }
  \end{minipage}
  \vspace{-6mm}
\end{table}

\vspace{-3mm}\paragraph{\textbf{Semi-supervised classification.}}
We evaluate the fine-tuning performance of the pretraind models for semi-supervised classification with 1\% and 10\% of labeled IN-1K samples, selected by SimCLR~\cite{chen2020simple}. We fine-tune the models for 20 epochs with the classifier learning rate 1.0 (0.2) and backbone learning rate 0.00001 (0.02) for 1\% (10\%) subset with a cosine-annealed scheduler. \cref{tab:exp_semi} presents the results of top-1 and top-5 accuracy on the validation set of IN-1K. Our method consistently outperforms the baseline approaches across different training durations. With 1\% labels, GenView pretrained for 200 epochs with MoCov2 achieves an improvement of +8.5\% in top-1 accuracy, and the one pretrained for 300 epochs with MoCov3 still improves top-1 accuracy by +1.9\%.

\vspace{-3mm}\paragraph{\textbf{Transfer learning on object detection and instance segmentation.}}
We evaluate the transfer learning performance of the pretrained models on MS-COCO object detection and instance segmentation benchmarks~\cite{lin2014microsoft}. The models are pretrained on IN-1K for 200 epochs, followed by fine-tuning on the $\operatorname{train2017}$ split and evaluation on the $\operatorname{val2017}$ split. We use a batch size of 16 and follow Detetron2's $1\times$ schedule~\cite{wu2019detectron2}, consisting of 90k training iterations with learning rate decay at the 60k-th and 80k-th iterations by a factor of 10. Both tasks utilize Mask R-CNN~\cite{he2017mask} with ResNet-50-FPN~\cite{lin2017feature} backbone. \cref{tab:exp_coco} presents the results of bounding box AP and instance mask AP. We observe that GenView is also able to enhance the downstream performances. When integrated on SimSiam, MoCov2, and BYOL, GenView excels in all metrics for detection and instance segmentation, highlighting its capacity to improve representation learning for complex localization and pixel-level tasks. Additionally, FreeATM also generates the same number of images as GenView using augmented prompts~\cite{zhang2023free}. We notice that GenView surpasses FreeATM on object detection even without relying on text prompts, emphasizing our approach's effectiveness.

\begin{table}[!t]
  \begin{minipage}{0.45\linewidth} 
\centering
  \caption{\textbf{Comparison with naive data augmentation methods under linear evaluation on IN-1K}. Models with ResNet-50 backbone are pretrained for 50 epochs on expanded datasets. The 4-th row incorporates 0.3M synthetic images produced by the generative model. The last row uses our framework in Sec. \ref{method:framework}. with only 0.15M synthetic images.}
  \label{tab:exp_naive}
  \scalebox{0.68}{
  \begin{tabular}{lccc}
  \toprule
    Dataset& Images& Top-1 & Top-5 \\ \midrule
    IN-1K& 1.28M& 62.39 & 84.57 \\
    IN-1K + Laion400M~\cite{schuhmann2022laion}& 1.28M + 0.3M& 63.31 & 85.53 \\
    IN-1K + ImageNet-21K~\cite{ridnik2021imagenet}& 1.28M + 0.3M& 64.10 & 85.86 \\
    IN-1K + Synthetic images & 1.28M + 0.3M & 63.36 & 85.14 \\
    \rowcolor{mygray} IN-1K + Our framework & 1.28M + 0.15M& \textbf{65.62} & \textbf{87.25} \\
    \bottomrule
  \end{tabular}
  }
  \end{minipage}
  \hfill
  \begin{minipage}{0.5\linewidth}
  \centering
  \caption{\textbf{Comparison with other view construction methods under linear evaluation on different datasets}. ResNet-18 is used as the backbone.}
  \vspace{-3mm}
  \label{tab:exp_tiny}
  \scalebox{0.75}{
    \begin{tabular}{lccc}
    \toprule
    Methods & CF10& CF100& TinyIN\\ \midrule
    \multicolumn{4}{c}{\emph{Variance within instance}}\\
    MoCov2 + C-Crop~\cite{peng2022crafting}& 88.78& 57.65& 47.98\\
    BYOL + C-Crop~\cite{peng2022crafting}& 92.54& 64.62 &47.23\\
    \midrule
    \multicolumn{4}{c}{\emph{Variance within pretraining datasets}}\\
    SimCLR + ViewMaker~\cite{tamkin2020viewmaker}& 86.30& -& -\\
    SimCLR + NTN~\cite{kim2023neural} & 86.90 & - & - \\
    MoCov2 + LMA~\cite{yang2022local} & 92.02 & 64.89 & - \\
    SimSiam + LMA~\cite{yang2022local} & 92.46 & 65.70 & - \\
    Simsiam + DiffAug ~\cite{zang2023boosting} & 87.30& 60.10& 45.30 \\
    \midrule
    \multicolumn{4}{c}{\emph{Variance beyond pretraining datasets}}\\
    W-perturb ~\cite{han2023constructive}& 92.90 & -& 51.05 \\
    \rowcolor{mygray} MoCov2 + GenView & 93.00 & 67.49 & \textbf{56.76}\\ 
    \rowcolor{mygray} BYOL + GenView & \textbf{93.56} & \textbf{67.53} & 54.79 \\ 
    \bottomrule
    \end{tabular}
  }
  \end{minipage}
  \vspace{-6mm}
\end{table}

\vspace{-3mm}\paragraph{\textbf{Comparison with naive augmentation methods.}}
We evaluate our method by comparing it to traditional data augmentation techniques. We extend IN-1K by incorporating 0.3 million images from Laion400M~\cite{schuhmann2021laion} and 0.3 million from ImageNet-21K~\cite{ridnik2021imagenet} (IN-21K). All experiments utilize MoCov3 with ResNet-50, which is pretrained for 50 epochs on these extended datasets. \cref{tab:exp_naive} presents the results of linear evaluation on IN-1K. Expanding IN-1K with Laion400M (2nd row) or synthetic images (4-th row) yields a slight improvement in top-1 accuracy, suggesting a limited contribution when directly incorporating images with domain gap. Extending IN-1K with IN-21K improves more than Laion400M, indicating the benefits from more training data in a similar domain. The most impressive results are obtained when using our framework with only 0.15 million generated images, leading to a remarkable 3.2\% improvement in top-1 accuracy, demonstrating that the effectiveness of our framework mainly stems from better pair construction, instead of introducing more training data. 

\vspace{-3mm}\paragraph{\textbf{Comparison with other view construction methods.}}
To evaluate GenView's effectiveness in enhancing SSL models compared to existing positive view construction methods, we conduct pretraining and evaluation on CIFAR-10\cite{krizhevsky2009learning} (CF10), CIFAR-100~\cite{krizhevsky2009learning} (CF100), and Tiny ImageNet~\cite{le2015tiny} (TinyIN) datasets. We train ResNet-18~\cite{he2016deep} for 500/500/200 epochs on CF10/CF100/TinyIN. For linear evaluation on validation sets of these datasets, the classifier is trained for 100 epochs using the SGD optimizer with a cosine-annealed learning rate of 0.2, no weight decay, and momentum of 0.9. As shown in \cref{tab:exp_tiny},
the methods are categorized based on the source of variance they use in data augmentation: within instance, within the pretraining datasets, and beyond the pretraining datasets. GenView, when combined with MoCov2, consistently outperforms the other data augmentation methods in SSL, demonstrating its effectiveness in borrowing rich knowledge from large-scale datasets to construct high-quality positive views.

\subsection{Ablations}
\label{exp:ablation}

\begin{table}[!t]
  \begin{minipage}{0.45\linewidth}
  \centering
 \caption{\textbf{Influence of each component under linear evaluation on IN-100.} ResNet-18 models are pretrained on IN-100 for 100 epochs. \textbf{Our framework} refers to using our framework to construct views but without dynamically adjusting the noise perturbation and the quality-driven contrastive loss. \textbf{Ada.View} represents our proposed adaptive view generation method. \textbf{Qual.Driv.Cont} indicates the use of our quality-driven contrastive loss.}
 \vspace{-3mm}
 
 \label{tab:exp:component}
 \scalebox{0.6}{
  \begin{tabular}{cccc}
  \toprule
  Our framework & Ada.View & Qual.Driv.Cont & Top-1 \\
  \midrule
  $\times$ & $\times$ & $\times$ & 65.52\\
  $\times$ & $\times$ & $\checkmark$ & 66.97 ($\uparrow$ 1.45)\\
  $\checkmark$ & $\times$ & $\times$ & 71.50 ($\uparrow$ 5.98)\\
  $\checkmark$ & $\checkmark$ & $\times$ & 73.96 ($\uparrow$ 8.44)\\
  $\checkmark$ & $\times$ & $\checkmark$ & 74.88 ($\uparrow$ 9.36)\\
  $\checkmark$ & $\checkmark$ & $\checkmark$ & \textbf{75.40 ($\uparrow$ 9.88)}\\
  \bottomrule
 \end{tabular}
\vspace{-5mm}
 }
    
  \end{minipage}
  \hfill
  \begin{minipage}{0.5\linewidth}
    \caption{\textbf{Influence of different noise level selection strategies under linear evaluation on IN-100.} ResNet-18 models are pretrained on IN-100 for 100 epochs.}
      \vspace{-2mm}
    \label{tab:exp_noiselevel}
    \scalebox{0.65}{
    \begin{tabular}{cccccccc}
        \toprule
        Method & CS(0) & CS(100) & CS(200) & CS(300) & CS(400) & RS & AS \\
        \midrule
        Top-1 &  71.80 &  72.14&  71.50 & 71.76 &  72.08 & 72.96 & \textbf{73.96}\\
        Top-5 &  92.19 & 92.34  & 91.88 & 92.02 & 92.36  & 92.78 &\textbf{93.22}\\
        \bottomrule
    \end{tabular}
    }

    \vspace{6mm}

    \centering
    \caption{\textbf{Influence of GenView application probability under linear classification on IN-1K.} Models with ResNet-50 backbone are pretrained for 50 epochs on IN-1K.}
    \vspace{-2mm}
\label{tab:exp_prob}
\scalebox{0.75}{
\begin{tabular}{ccccccc}
  \toprule
   $\alpha$& 0& 0.1& 0.3& 0.5& 0.8& 1.0\\
  \midrule
  Top-1 & 62.39 & 65.86 & 68.38 & 69.04 & 69.47& \textbf{70.55} \\
  Top-5 & 84.57 & 87.10 & 89.02 & 89.29 & 89.49& \textbf{90.34} \\
  \bottomrule
\end{tabular}
    }
  \end{minipage}
  \vspace{-6mm}
\end{table}

\paragraph{\textbf{Influence of each component.}}
We evaluate the contributions of individual components as well as their combinations. ResNet-18 models are pretrained on IN-100 for 100 epochs using MoCov3 as the baseline, with a batch size of 512. IN-100 is a subset of IN-1K selected by \cite{tian2019contrastive}. For conditioning the generation of positive views with GenView, we employ 50,000 randomly selected class-balanced images from IN-100. We use a cosine decay learning rate schedule and employ the LARS optimizer with a learning rate of 1.2, weight decay of 1e-6, and momentum of 0.9. Linear evaluation settings are consistent with those detailed in \cref{tab:exp_linear}, with a training duration of 50 epochs.
\cref{tab:exp:component} offers valuable insights:
(1) Utilizing our framework but without our adaptive view generation significantly enhances accuracy, achieving a top-1 accuracy improvement of 5.98\% compared to the baseline.
(2) The incorporation of adaptive view generation further elevates model performance, resulting in an improvement of 8.44\% (from 65.52\% to 73.96\%).
(3) The quality-driven contrastive loss also plays a pivotal role in our framework. It can further improve the performance of adaptive view generation. Applying the quality-driven contrastive loss to the baseline method leads to a modest gain of 1.45\% (from 65.52\% to 66.97\%). However, when combined with our framework, a more substantial performance improvement of 3.38\% (from 71.50\% to 74.88\%) is observed. This highlights the effectiveness of our framework and also the importance of the proposed modules in enhancing contrastive learning by improving the quality of positive pairs.

\vspace{-3mm}\paragraph{\textbf{Influence of the noise level selection strategies.}}
We examine the impact of different noise level selection strategies on SSL performance in \cref{tab:exp_noiselevel}. Three strategies are compared: Constant Selection (CS), Random Selection (RS), and Adaptive Selection (AS). CS applies a uniform noise level $c$ to all samples, with experiments conducted at various fixed levels (CS(0), CS(100), CS(200), CS(300), CS(400)). RS introduces variability by randomly selecting noise levels from the set ${0, 100, 200, 300, 400}$. AS dynamically adjusts noise levels based on the input image's foreground proportion, as guided by Eq.~(\ref{eq:l_ada}). 
We use the same pretraining and linear evaluation settings as \cref{tab:exp:component}. The results indicate that AS achieves the highest accuracy at 73.96\%, demonstrating the advantage of dynamically adjusting noise levels according to input characteristics. CS and RS yield lower performance, because static or random noise levels may result in overly similar or false positive pairs.

\vspace{-3mm}\paragraph{\textbf{Influence of the probability to apply GenView.}}
The impact of different probabilities ($\alpha$) for applying GenView augmentation is shown in \cref{tab:exp_prob}. An increase of the probability ($\alpha$) of applying GenView results in improved model performance, with top-1 accuracy consistently increasing from 62.39\% at $\alpha=0$ to 70.55\% at $\alpha=1.0$. This highlights the significance of a higher GenView application probability in enhancing the model's ability to learn meaningful representations. By default, we set $\alpha=1$ for all the experiments in our main results.

\vspace{-4mm}
\subsection{Qualitative Evaluation}
\label{exp:vis}
\vspace{-1mm}

\begin{figure}[t]
  \centering
  \includegraphics[width=0.75\linewidth]{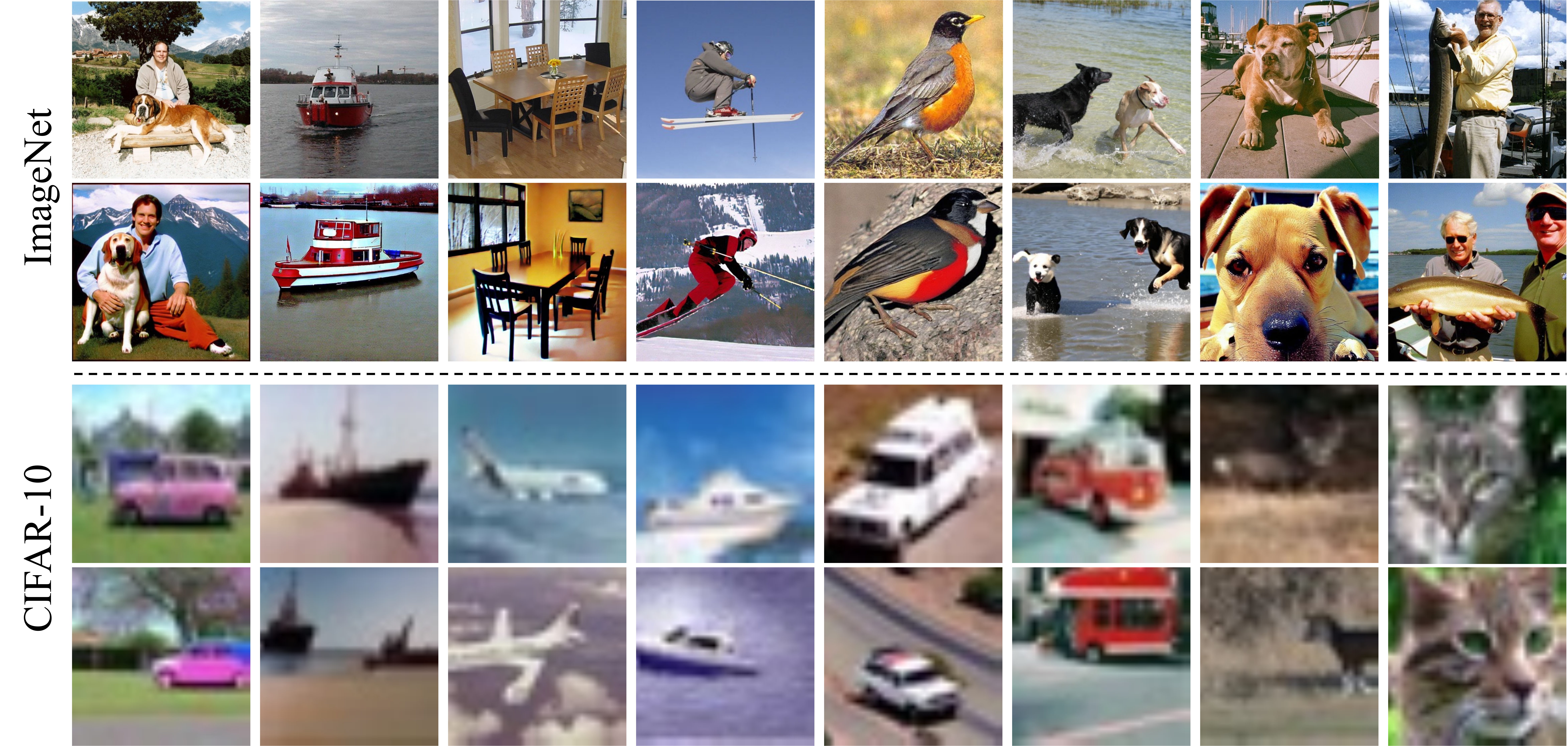}
  \vspace{-2mm}
  \caption{The positive pair of views constructed by GenView conditioned on images from IN-1K, and CF10.}
  \label{fig:vis_genview}
  \vspace{-6mm}
\end{figure}
A qualitative illustration of the positive views constructed by GenView is shown in \cref{fig:vis_genview}. The top rows display original images, and the bottom rows show images generated by GenView. This visualization demonstrates GenView's capacity to introduce variations in background, pose, and view angle while preserving the main semantics, which is crucial for learning invariant representations. More visual examples are provided in the \cref{subb_vis}.

%% file: sec/5_conclusion.tex
\vspace{-4mm}
\section{Conclusion}
\vspace{-2mm}
In this paper, we aim to address the challenge of creating diverse and semantically coherent positive views for SSL. We introduce GenView, a framework that leverages the ability of pretrained generative model in a controllable way to enhance the view quality. It employs an adaptive view generation method that dynamically adjusts noise levels for controlled variability. The quality-driven contrastive loss prioritizes high-quality positive pairs with greater foreground similarity and background diversity while diminishing the impact of low-quality or even false pairs. 
Experiments demonstrate that GenView consistently improves the SSL performance in various tasks, and outperforms other view augmentation methods. Ablation studies analyze the efficacy of each component, and qualitative evaluation shows its effectiveness in constructing views with background, pose, and view angle variations. 

\vspace{-4mm}
\section*{Acknowledgements}
\vspace{-2mm}
This work was supported in part by the National Natural Science Foundation of China under Grant 62376069, in part by Young Elite Scientists Sponsorship Program by CAST under Grant 2023QNRC001, and in part by Guangdong Basic and Applied Basic Research Foundation under Grant 2024A1515012027. The work was also supported by funding from KAUST Center of Excellence on GenAI, under award number 5940.

%% file: sec/X_suppl.tex
\clearpage
\appendix
\setcounter{page}{1}
\section{Implementation Details}
\label{supp_details}
\vspace{-2mm} \paragraph{\textbf{Method for adding noise perturbations.}}
We use the pretrained Stable unCLIP v2-1\footnote{\url{https://huggingface.co/stabilityai/stable-diffusion-2-1-unclip}} model to generate image variations based on CLIP image embeddings $c$. An empty string serves as the text prompt to avoid any reference to image contexts or object names. The noised image embedding with perturbation strength $l$ is defined as: $\text{noisy}(\textbf{c},l) = \sqrt{\overline{a}_l}\mathbf{c} + \sqrt{(1-\overline{a}_l)} \varepsilon$, where $\varepsilon \sim \mathcal{N}(0, \mathbf{I})$, and $\overline{a}_l$ is the cumulative product of $\alpha_i$ values for $i$ ranging from 0 to $l$. Each $\alpha_i$ is defined as $1-\beta_i$, with $\beta_i$ representing the noise variance introduced at step $i$, following the default linear schedule for $\beta_{[1:l]}$ from DDPM~\cite{ho2020denoising}. Higher values of the perturbation strength $l$ result in increased diversity in the generated images.

\vspace{-2mm} \paragraph{\textbf{Method for calculating foreground proportion.}}
We use the pretrained CLIP ViT-H/14 backbone\footnote{\url{https://huggingface.co/laion/CLIP-ViT-H-14-laion2B-s32B-b79K}}, which serves as the conditional image encoder in Stable UnCLIP v2-1, for the encoder $C$ used in determining the proportion of foreground content before image generation. This backbone generates 256 tokens with a dimension of 1280 from a $224^2$ input resolution. For calculating PCA features, 10,000 images are randomly sampled from the original dataset. The threshold $a$ in \cref{eq:proportion} is selected to ensure that foreground tokens account for approximately 40\% of the total tokens, providing a clear separation between foreground and background as depicted in the ~\cref{method:adaptive}.

\vspace{-2mm} \paragraph{\textbf{Method for generating attention maps.}}
We employ the pretrained CLIP ConvNext-Base (with wide embedding dimension) backbone\footnote{\url{https://huggingface.co/laion/CLIP-convnext_base_w-laion2B-s13B-b82K-augreg}} as the encoder $E$ to extract feature maps from augmented positive views. These feature maps have a resolution of $7^2$ based on a $224^2$ input resolution. We compute foreground and background attention maps using the PCA vector computation method described in the previous paragraph.

\vspace{-2mm} \paragraph{\textbf{Hyper-parameters for view generation.}}
We generate one augmented image for each image in the training set of IN-1K/CF10/CF100/TinyIN, with $T$ (the number of denoising steps) set to 20 for efficiency. The classifier-free guidance scale~\cite{ho2022classifier} is set to 10 to ensure image quality. The diversity of generated images is controlled by the level of noise perturbations applied to the image embeddings. To match the original dataset sizes of IN-1K/CF10/CF100/TinyIN, we resize the generated images from their original resolution of $768^2$ to $512^2$/$32^2$/$32^2$/$64^2$, respectively.

\vspace{-2mm} \paragraph{\textbf{Comparison with SSL methods.}}
\begin{table}[t]
    \centering
    \caption{Hyperparameters for comparison with SSL models pretrained on IN-1K.}
    \vspace{-3mm}
    \resizebox{1\linewidth}{!}{
    \begin{tabular}{lcccccc}
    \toprule
                  & MoCov2 & SwAV & SimSiam  & BYOL&MoCov3 &MoCov3 \\ \midrule
    Optimizer     & SGD     & LARS& SGD  & LARS&LARS&AdamW\\
    Learning Rate & 0.03& 0.6& 0.05& 4.8&1.2/9.6/4.8&2.4e-3\\
    Weight Decay  & 1e-4& 1e-4  & 1e-4  & 1e-6&1e-6&0.1\\
    Momentum      & 0.9     & 0.9  & 0.9  & 0.9      &0.9       &-\\
    Cosine Decay  & \checkmark & \checkmark  & \checkmark  & \checkmark &\checkmark  &\checkmark  \\
    Batch Size    & 256& 256& 256& 4096&512/4096/4096&4096/4096\\
    Loss& $\mathcal{L}_{SSL, NCE}$& $\mathcal{L}_{SSL, KL}$& $\mathcal{L}_{SSL, COS}$& $\mathcal{L}_{SSL, COS}$&$\mathcal{L}_{SSL, NCE}$&$\mathcal{L}_{SSL, NCE}$\\
    Epochs        & 200& 200& 200& 200&100/300&300\\
    Backbone      & ResNet50& ResNet50& ResNet50& ResNet50 &ResNet50  &VIT-S/ViT-B\\
    Embedding Dim & 2048& 2048& 2048& 2048      &2048&384/768\\
    Projection Dim & 128& 128& 2048& 256&256&256\\
    \bottomrule
    \end{tabular}
    }
    \label{tab:exp_hyperparam}
    \vspace{-4mm}
\end{table}
The baseline results of MoCov3 in \cref{tab:exp_linear} are from the public codebase\footnote{\url{https://github.com/facebookresearch/moco-v3}}. Hyper-parameters for comparison with other SSL methods pretrained on IN-1K are listed in \cref{tab:exp_hyperparam}.

\vspace{-2mm} \paragraph{\textbf{Comparison with naive augmentation methods.}}
To expand IN-1K with additional training data without introducing new classes, we employ a retrieval-based technique~\cite{beaumont-2022-clip-retrieval} for expanding IN-1K with Laion400M. We query the entire Laion400M dataset with 0.3 million randomly sampled IN-1K images and select the most similar image for each query image. For expanding IN-1K with IN-21K, we randomly sample 0.3 million non-repeating images with labels matching those in the IN-1K dataset. For GenView, positive views are generated for 0.15 million randomly sampled IN-1K images. For the experiments, the ResNet-50 models are pretrained on the expanded dataset using a batch size of 512. We apply a cosine decay learning rate schedule and use the LARS optimizer with a learning rate of 1.2, weight decay of 1e-6, and momentum of 0.9. Linear evaluation settings align with those in \cref{tab:exp_linear}.

\vspace{-2mm} \paragraph{\textbf{Comparison with other view construction methods.}}
We compare our approach with several baseline methods, including ContrastiveCrop~\cite{peng2022crafting} (C-Crop), ViewMaker~\cite{tamkin2020viewmaker}, Neural Transform Network~\cite{kim2023neural} (NTN), Local Manifold Augmentation method~\cite{yang2022local} (LMA), Diffusion-based augmentation from scratch~\cite{zang2023boosting} (DiffAug), and $\mathcal{W}$-perturb~\cite{han2023constructive}. For our experiments, we use three datasets: CF10, CF100, and TinyIN. We employ SGD as the optimizer with a learning rate of 0.5, weight decay of 5e-4, and momentum of 0.9. The learning rate follows a linear warm-up for 10 epochs and then switches to the cosine decay scheduler. Batch sizes are set to 512 for CF10 and CF100 and 256 for TinyIN. The momentum coefficient for the momentum-updated encoder and memory buffer size is set to 0.99/0.99/0.996 and 4096/4096/16384 for CF10/CF100/TinyIN, respectively. We use the $\mathcal{L}_\text{SSL, NCE}$ loss with a temperature of 0.2 and train for 500 epochs on CF10 and CF100 and 200 epochs on TinyIN. The backbone architecture used is ResNet18 with an embedding dimension of 512 and a projection dimension of 128. We replace the first 7x7 Conv of stride 2 with 3x3 Conv of stride 1 and remove the first max-pooling operation. For data augmentations, we use random resized crops (the lower bound of random crop ratio is set to 0.2), color distortion (strength=0.4) with a probability of 0.8, and Gaussian blur with a probability of 0.5. The images from the CF10/CF100 and TinyIN datasets are resized to 32x32 and 64x64 resolution.

\section{Additional Illustration}
\label{subb_vis}
\noindent \textbf{Further visualization and limitation analysis.}
\begin{figure*}[t]
  \centering
  \includegraphics[width=0.8\linewidth]{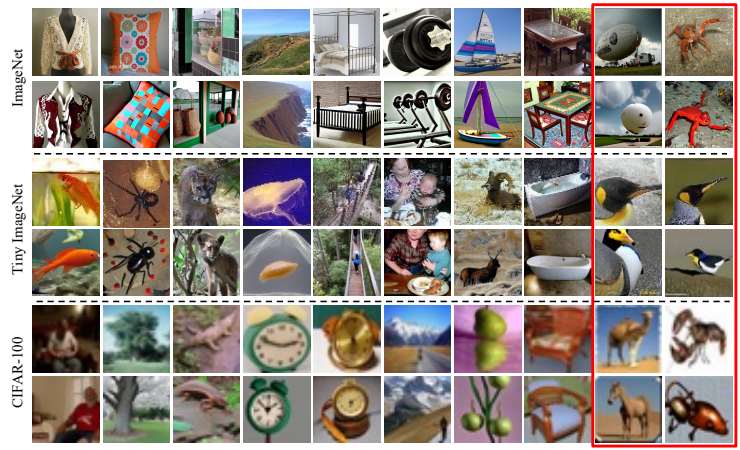}
  \caption{Visualization of positive pairs generated by GenView, depicting successful variations and failure cases (outlined in red).}
  \label{fig:vis_supp}
\end{figure*}
\cref{fig:vis_supp} displays positive pairs generated by GenView, showing its ability to introduce variations while maintaining semantic consistency. Notable observations include:

\begin{itemize}
    \item Capacity constraints in the CLIP conditional encoder or generator can challenge the accurate representation of long-tailed categories or complex scenes, resulting in less realistic generations, such as the generated airships (2nd row of the 9th column) and lobsters (2nd row of the 10th column).
    \item The granularity of conditional images is crucial, as lower-resolution images can lead to a loss of detail and misclassification of the generated images. For instance, conditioning on a camel image in the 5th row of the 9th column with a resolution of $32^2$ produces a generated image resembling a horse, losing the camel's distinctive features.
    \item Partially visible objects, like the head of the king penguin in the 3rd row of the 9th column, may result in generation errors, yielding images resembling ducks (4th row of the 9th column) or birds (4th row of the 10th column).
\end{itemize}

Despite these limitations, GenView's adaptive view generation method ensures that the synthesized samples maintain attributes similar to the conditional images, providing valuable information for SSL training. Additionally, our quality-driven contrastive loss mechanism addresses semantic inconsistencies, mitigating their impact on contrastive representation learning. Future work will focus on refining diffusion models to enhance generative augmentation and address the highlighted failure cases.

\begin{table}
\centering
\vspace{-2mm}
\caption{Average cosine similarity between positive views and original images. \textbf{Retrieval} refers to pairs constructed by retrieving the most similar image from Laion400M and pairing it with the query image. \textbf{RS} and \textbf{AS} refer to methods in \cref{tab:exp_noiselevel}.}
\label{tab:cosine_similarity}
\scalebox{0.9}{
\begin{tabular}{lccc}
\toprule
Method & Retrieval& RS&AS (ours)\\
\midrule
Cosine Similarity& 0.674  & 0.729 & \textbf{0.743}\\
\bottomrule
\end{tabular}
}
\vspace{-5mm}
\end{table}

\vspace{-2mm} \paragraph{\textbf{Evaluation of positive views constructed by different methods.}}
We calculate the average cosine similarity between the original images and their associated positive views for different methods. We randomly sample 50,000 images from IN-1K and compute the cosine similarity of CLIP image embeddings for each pair. The results are presented in \cref{tab:cosine_similarity}, which demonstrates that our method produces positive views with significantly higher semantic similarity to the original images compared to the RS and Retrieval Laion400M methods.

\section{Algorithm}
\label{supp_algo}
The GenView algorithm is detailed in Algorithm~\ref{alg:genview}.

\begin{algorithm}[h]
\caption{GenView}
\label{alg:genview}
\begin{algorithmic}[1]
\Require Original images $\mathbf{X}_{1:n}$, CLIP image encoder $C(\cdot)$, pretrained diffusion model $\mathcal{G}(\cdot)$

\Statex \textbf{Offline Adaptive View Generation}
\For{each image $\mathbf{X}_i$ in $\mathbf{X}_{1:n}$}
    \State $\mathbf{Z}_i, \mathbf{c}_i \gets C(\mathbf{X}_i)$
    
    \State $\mathbf{A}_i \gets \text{Normalize}(\text{PCA}(\mathbf{Z}_i))$ 
    
    \State ${p}_i \gets \frac{\sum B(\mathbf{A}_i, a)}{H \times W}$ 
    
    \State $l^\text{ada}_i \gets \mathcal{F}^{\text{ada}}({p}_i)$ 
    
    \State $\mathbf{X}^+_i \gets \mathcal{G}(\mathcal{N}(0, \mathbf{I}), \text{noisy}(\mathbf{c}_i, l^\text{ada}_i), w)$ 
\EndFor

\Statex \textbf{Training with Quality-driven Contrastive Loss}
\For{each image $\mathbf{X}_i$ and its corresponding $\mathbf{X}_i^+$}
    \State $\mathbf{P}_i, \mathbf{P}_i^+ \gets t(\mathbf{X}_i), t^+(\mathbf{X}_i^+)$ 
    
    \State $\mathbf{F}_i, \mathbf{F}_i^+ \gets E(\mathbf{P}_i), E(\mathbf{P}_i^+)$ 
    
    \State Compute $\mathbf{M}_i^f, \mathbf{M}_i^{f+}, \mathbf{M}_i^b, \mathbf{M}_i^{b+}$ 
    
    \State Compute $\mathbf{z}_i^{f}, \mathbf{z}_i^{f+}, \mathbf{z}_i^{b}, \mathbf{z}_i^{b+}$ 
    
    \State $s_i^f, s_i^b \gets \text{sim}(\mathbf{z}_i^{f}, \mathbf{z}_i^{f+}), \text{sim}(\mathbf{z}_i^{b}, \mathbf{z}_i^{b+})$
    
    \State $q_i \gets s_i^f - s_i^b$ 
    
    \State $w_i \gets \text{Softmax}(q_i)$ 
    
    \State $\mathcal{L}_\text{SSL,*} \gets \text{ContrastiveLoss}(\mathbf{P}_i, \mathbf{P}_i^+)$
    
    \State $\tilde{\mathcal{L}}_\text{SSL,*} \gets w_i \mathcal{L}_\text{SSL,*}$ 
\EndFor
\end{algorithmic}
\end{algorithm}